\documentclass{article}

     \usepackage[final,nonatbib]{nips_2018}

\usepackage[utf8]{inputenc} % allow utf-8 input
\usepackage[T1]{fontenc}    % use 8-bit T1 fonts
\usepackage{hyperref}       % hyperlinks
\usepackage{url}            % simple URL typesetting
\usepackage{booktabs}       % professional-quality tables
\usepackage{amsfonts}       % blackboard math symbols
\usepackage{nicefrac}       % compact symbols for 1/2, etc.
\usepackage{microtype}      % microtypography
\usepackage{graphicx}
\usepackage{amsmath}
\usepackage[font=small]{caption}

\title{  Secondary Protein Structure Prediction Using Neural Networks }

\author{%
  Sidharth Malhotra \\
  Department of Computer Science\\
  Northeastern University \\
  360 Huntington Ave.,
  Boston, MA 02115 \\
  \And
  Robin Walters \\
  Department of Mathematics\\
  Northeastern University \\
  360 Huntington Ave.,
  Boston, MA 02115 \\
}

\begin{document}

\maketitle

\vspace{-0.2in}

\begin{abstract}
  In this paper we experiment with using neural network structures to predict a protein's secondary structure ($\alpha$-helix positions) from only its primary structure (amino acid sequence).  We implement a fully connected neural network (FCNN) and preform three experiments using that FCNN.  Firstly, we do a cross-species comparison of models trained and tested on mouse and human datasets.  Secondly, we test the impact of varying the length of protein sequence we input into the model.  Thirdly, we compare custom error functions designed to focus on the center of the input window.  At the end of paper we propose a alternative, recurrent neural network model which can be applied to the problem. 
\end{abstract}

\vspace{-0.2in}

\section{Introduction}

\vspace{-0.1in}
\subsection{Background: Protein Structure}
\vspace{-0.1in}

Proteins are large molecules in living cells which perform many different roles, for example, catalyzing biochemical reactions, signaling across the cell, and transmitting and verifying genetic information. This makes them one of the most essential molecules for living organisms \cite{jiang2017protein}. 

Chemically, proteins are formed by long chains of small molecules called amino acids bonded together.  There are 20 different kinds of amino acids commonly found in eukaryotes.  The amino acid sequence is the protein's \emph{primary structure}.  The chemical and physical properties of the different amino acids, together with the flexibility of the bonds between them, cause the long chain to fold into various shapes, the most common of which are $\alpha$-helices, $\beta$-sheets and $\beta$-turns.  These shapes constitute the protein \emph{secondary structure}.  In turn, these secondary structures fold and refold into more complex 3D shapes, the \emph{tertiary structure} of the protein.  Lastly, when two or more of these polypeptide chains fold together, they form a \emph{quaternary} structure \cite{wiki1, partsci}. 

%Functionality of a protein is closely related its structure, therefore, making it very important to predict the 3D structure of a protein with high accuracy. What makes the problem of predicting protein structures interesting and complex at the same time is the fact that the percent of structures known among all known protein sequences is only about 0.2\%.  The recognized functions are even less. This poses a big challenge for biologists to understand the structures and functions of proteins \cite{jiang2017protein}.

%Folded protein structure is categorized into four levels - primary, secondary, tertiary and quaternary \cite{wiki1}. The primary structure refers to the specific sequence of amino acids. The secondary structure refers to the compact rotations of amino acids around bonds within a protein. These rotations make proteins flexible and foldable into wide variety of shapes. The three most important secondary structures are alpha-helix, beta-sheets and beta-turns. The tertiary structure refers to the folds and refolds by polypeptide to form a complex 3D shape. The quaternary structures are composed of two or more polypeptide chains \cite{partsci}. 

%The 3D structure and functionalities of a protein are determined by primary and tertiary structures whereas the secondary structures bridges both primary and tertiary structures. This makes the study of secondary protein structures a necessary and important milestone to predict the 3D structure of a protein which can further allow us to understand the complex functionalities of a protein \cite{yoo2008machine}.

\vspace{-0.1in}
\subsection{Problem}
\vspace{-0.1in}

The structure of a protein determines its functionality within the cell and how it interacts with other molecules such as drugs.  It is thus very important to determine the 3D structure of a protein with high accuracy.  Learning the primary structure, or sequence, of a protein is relatively cheap and easy, but finding the full folded structure is expensive and difficult, requiring lab techniques such a nuclear magnetic resonance imaging and X-ray diffraction crystallography.  Consequently, the percent of structures known among all known protein sequences is only about 0.2\%.  The recognized functions are even less. This poses a big challenge for biologists trying to understand the structures and functions of proteins \cite{jiang2017protein}.  

In this paper, we address the problem of predicting folded structure using only the sequence, with the aim of providing a faster and cheaper alternative to lab-based methods.  Specifically, the problem we are trying to solve in this paper is as follows:

\textbf{Problem:}  \emph{Given as input only a protein's primary structure, its amino acid sequence, predict part of its secondary structure, which acids in the sequence are parts of an $\alpha$-helix.} 

The ultimate goal is to be able to predict the complete secondary, tertiary, and quaternary structure from the sequence alone, but this is currently out of reach.  Nonetheless, the study of secondary protein structures is considered a necessary and important milestone to predict the 3D structure of a protein and can further allow us to understand the complex functionalities of a protein \cite{yoo2008machine}.

\vspace{-0.1in}
\subsection{Challenges}
\vspace{-0.1in}

This problem is hard for several reasons.  Protein sequence lengths can be over 1000 amino acids long and with 20 different possible acids per position, the number of sequences is enormous.  This necessitates some method to limit input size such as using only small sections of the sequence, \emph{windows}, at a time.  Moreover, protein structure depends subtly on sequence.  Changing or mutating even a single amino acid can change the overall structure dramatically.  Lastly, \textit{a priori}, acids distant from each other in the sequence can still interact in the folded structure, making it difficult to limit the length of window too greatly.

\vspace{-0.1in}
\subsection{Contributions}
\vspace{-0.1in}

In this paper, we implement and train different machine learning models to predict the secondary structures of proteins. We compare and discuss the results from
%ensemble models like Random Forests Classifiers,
 deep learning models like fully connected neural networks (FCNN) and sequential models like recurrent neural networks (RNN).  We also perform three experiments using our models to help with protein structure classification.  Here is a summary of our contributions to the problem:

\begin{itemize}
    \item Cleaned and removed redundancy from a mouse and human protein dataset.  Built code to import PDB files into PyTorch. 
    \item Built and trained two models, a fully-connected neural network (FCNN), and a recurrent neural network (RNN).  The FCNN has a good prediction root mean squared error (RMSE) of 0.22.  
    \item Performed a cross-species comparison by training two FCNNs, one on a mouse protein dataset and one on a human protein dataset.  We then tested both models on both the mouse and human test sets.  The models performed best on their own species test sets, but did very well on the cross-species test as well.  This validates the concept of using mice as model organisms within this context. 
    \item Tested different window sizes of 7, 10, and 13 amino acids.  Despite our hypothesis that 10 or 13 would perform best, 7 had the lowest loss.  The implication is that at least for $\alpha$-helix prediction, sequence length and input size are not large challenges.
    \item Tested different error functions such as Gaussian, unweighted RMSE, and centered.  We found unweighted and Gaussian performed best.
\end{itemize}

\section{Related Work}
Yang \cite{yang2008protein} compared algorithmic efficiencies and computational times 
for single sequence protein prediction through neural network and 
support vector machines based algorithms. Yang tried different window 
and hidden layer sizes from the range 1 to 21 and 0 to 125 respectively. 
The neural network approach peaked with a performance of $67.42\%$ 
accuracy at a window size of 15 amino acids and a hidden layer of 75 units. 
Although, SVMs performed better at convergence than neural networks and did not 
tend to overfit, the overall performance of neural network outperformed that of the 
SVMs. 

Malekpour \cite{malekpour_naghizadeh_pezeshk_sadeghi_eslahchi_2009} 
improved the existing method of Segmental semi Markov models 
(SSMMs) using three neural networks that used multiple sequence alignment profiles.
The outputs of the neural networks were passed through SSMMs to predict secondary 
structures of amino acids. The proposed model predicted the protein structures with 
an overall accuracy of $75.35\%$. 

Jones \cite{jones_1999} attempted a PSIPRED method that achieved an accuracy score of between $76.5\%$ to $78.3\%$. 
PSIPRED works on position specific scoring matrices. The method is split into 
three main stages: generation of sequence profile, prediction of initial secondary 
structure, and filtering of the predicted structure.

King \cite{king1990machine} mentions about PROMIS, a machine learning program that predicted secondary structures in protein up to the accuracy of 60\%, using the generalized rules that characterize the relationship between primary and secondary structure in globular proteins.
\vspace{-.1in}

Pollastri \cite{pollastri2002improving} uses an ensemble of bidirectional RNNs to predict the protein secondary structures in three and eight categories resulting in deriving two new predictors. The predictors tested on three different test sets perform with an accuracy of 78\%.

\section{Method/Model} \label{model}
\vspace{-.1in}

\subsection{Feature Extraction and Engineering}
 The protein data bank (pdb) files that we selected for our dataset contained much information about each protein including the complete 3D structure of the protein, the amino acid sequence, and the location of various secondary structure such as $\alpha$-helices.  We extracted
helix structure information from all the pdb files and then used one-hot encoding to represent each amino acid type,
and binary values (1 or 0) to represent our target variables (helix structures) in a protein sequence.
There can be 20 different types of amino acids in a protein sequence which resulted in one-hot encoded
vectors of length 20 each. \cite{wiki2}  
 
\vspace{-.1in}

\subsection{Fully Connected Model}
We selected a fully connected neural network (FCNN) with 200 input neurons, 
40 hidden neurons, and 10 output neurons as our first training model.  This type of model is also called a multilayer perceptron (MLP).   The input vector represents a window of 10 sequential amino acids in the protein sequence.  We encode each type of amino acid using a standard basis vector $\mathbf{e}_i$ in $[0,1]^{20}$ and then flatten this into an input vector $\mathbf{x}$ of length 200.  Define the activation function $\mathrm{RELU}$ by
\[
\mathrm{R}(x) =
\begin{cases}
x, & \text{if } x \geq 0, \\
0, & \text{if } x < 0.
\end{cases}
\]
Then define the hidden layer values as
\[
\mathbf{h}_i = \mathrm{R}\left( \sum_{j=1}^{200} w^{(1)}_{i,j} \mathbf{x}_j \right)
\]
where $w^{(k)}_{i,j}$ are the models trainable weights.

The output vector is then defined 
\[
\mathbf{o}_i = \mathrm{R}\left( \sum_{j=1}^{40} w^{(2)}_{i,j} \mathbf{h}_j \right).
\]

We interpret these ten values as the predicted probability of finding a helix (or \emph{helicity}) at each of the 10 positions within the window.  We can reconstruct a prediction for the entire protein by moving this window over the entire sequence and extracting the middle values (or as close as possible) at each position. 

\begin{figure}[h]
\label{diagramofnet}
\begin{center}
\includegraphics[scale=0.495]{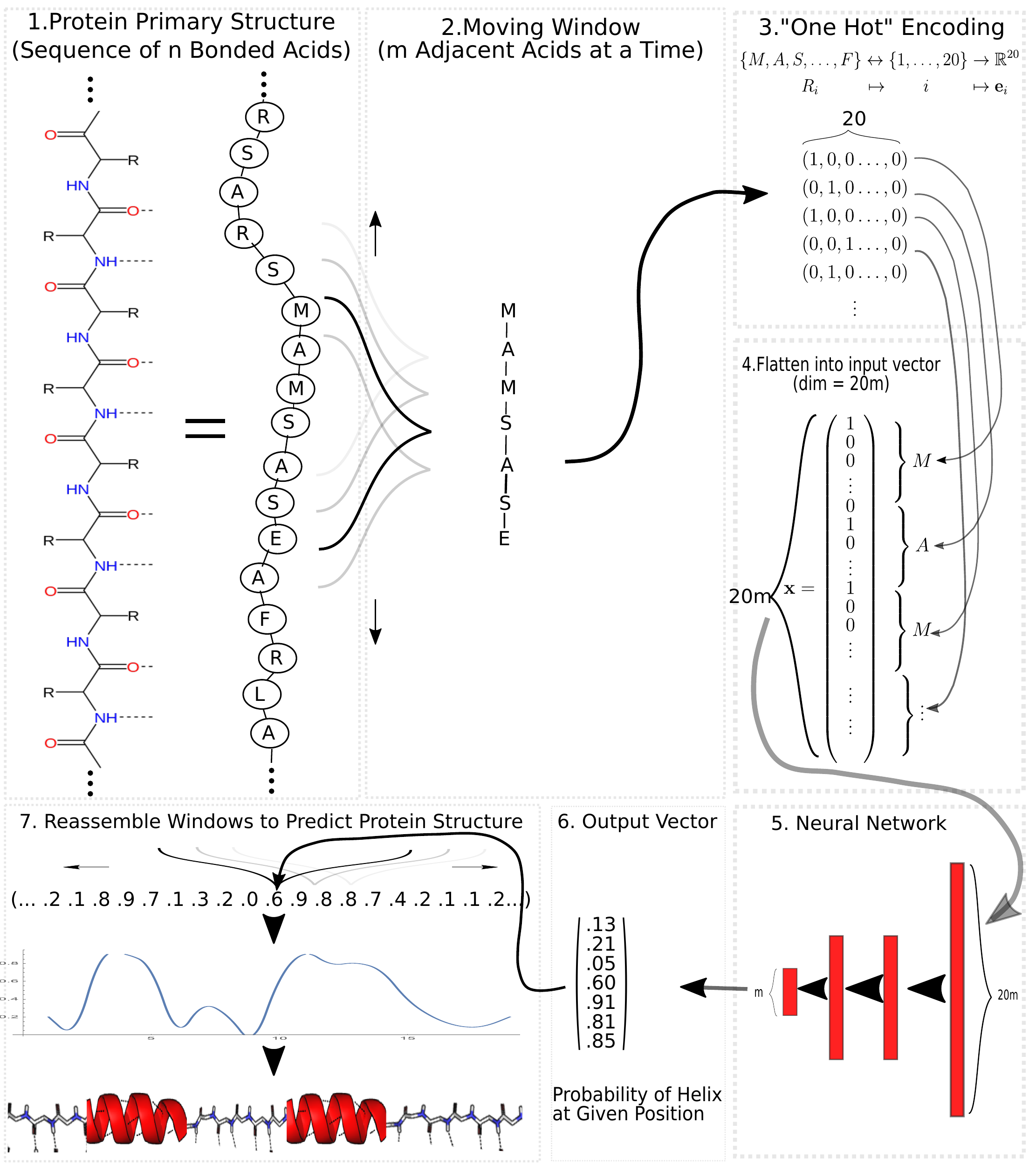}
\end{center}
\caption{Diagram of the FCNN.}
\label{fig-FCNN}
\end{figure}

Such a model can be thought of as analogous to a one-dimensional convolutional model in which the input and output vectors stretch over the entire protein sequence since we use the same shared weights across the entire sequence.  \autoref{fig-FCNN} is a diagrammatic version of our model.  

\vspace{-.1in}
%\newpage

%Note that while a fully connected network is shown in step 5, we plan to vary the type of network we put here.

\section{Experiment}
\subsection{Training Error Functions} \label{sec-setup}
\begin{figure}[ht]
\vspace{-0.1in}
\begin{center}
\includegraphics[width=0.75\textwidth]{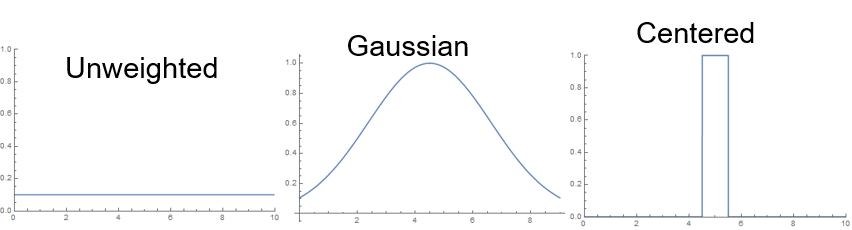}
\caption{Different Loss functions}
\label{fig-error_func}
\end{center}
\vspace{-0.2in}
\end{figure}

For the purpose of this paper, we implemented a total of three error functions, namely, unweighted loss, Gaussian loss and centered loss (See  \autoref{fig-error_func}).
All loss functions are weighted RMSE functions defined as such
\[
E(\overline{p},\overline{y}) = \sqrt{\sum_{i=1}^m w_i (p_i - y_i)^2},
\]
where $i$ indexes over the window of length $m$, $\overline{y}$ are the true values,  $\overline{p}$ are the predicted values and $\overline{w}$ is the weight vector.  

The unweighted error function uses $w_i = 1/m$ and is equivalent to root mean squared error (RMSE).  The Gaussian weight is defined
\begin{align}
u_i &= e^{\left(\frac{\left(i - m  \right)^2}{5 m}\right)^{4/3}} \\
w_i &= u_i / W
\end{align}
where $W = \sum_{i=1}^m u_i$.  The centered loss is defined
\begin{align}
w_i = 
\begin{cases}
1 - \frac{m-1}{100}, & \text{if } i = \left\lfloor \frac{m}{2} \right\rfloor  \\
\frac{1}{100},  & \text{ otherwise.}
\end{cases}
\end{align}

 In the unweighted loss function, all the acids in a single window of protein sequence have equal weight, whereas in Gaussian loss, the weight is greatest for the acid at the center of the window and decreases gradually towards the ends. At the extreme, for centered loss, only the acid at the center has significant weight whereas all other acids in the window have negligible weight. We selected these functions to test our hypothesis that in a moving window parsing of a protein sequence, acids at the center of the window should play a major role in predicting the accurate location of the helix structure in the protein sequence and should have better predictions.

\subsection{Test Criterion}

The above error functions are for training the models.  They are give the error between the true helicity and predicted helicity on a short window of the sequence.  For testing, it is necessary to use a standard error function to compare the results.  We use standard RMSE \textit{over the length of the entire protein} as our test criterion.  In this case, the predicted values of helicity are reconstructed by moving the window over the entire protein sequence and extracting a helicity predictions for each position using the window in which it is closest to the center.  \textit{Any references to average test loss throughout refer to this definition.}

\subsection{Dataset}
\begin{figure}[h]
  \centering
  \vspace{-0.3in}
  \begin{minipage}{0.45\textwidth}
      \centering
      \includegraphics[scale=0.25,angle=270]{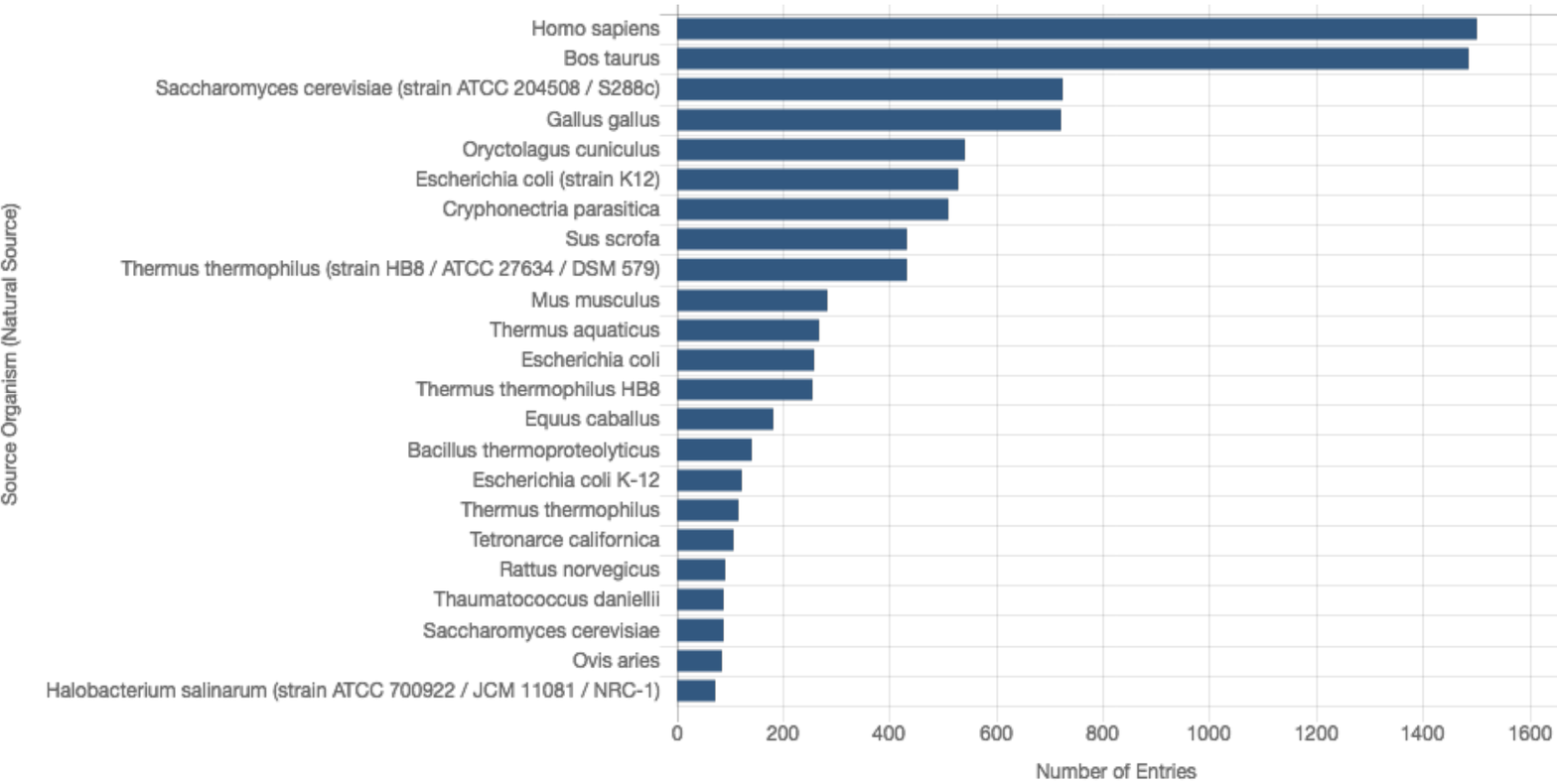} % first figure itself
       \vspace{-0.3in}
      \caption{Number of proteins in PDB by species.}
      \label{fig2}
  \end{minipage}\hfill
  \begin{minipage}{0.45\textwidth}
      \centering
      \includegraphics[scale=0.25,angle=270]{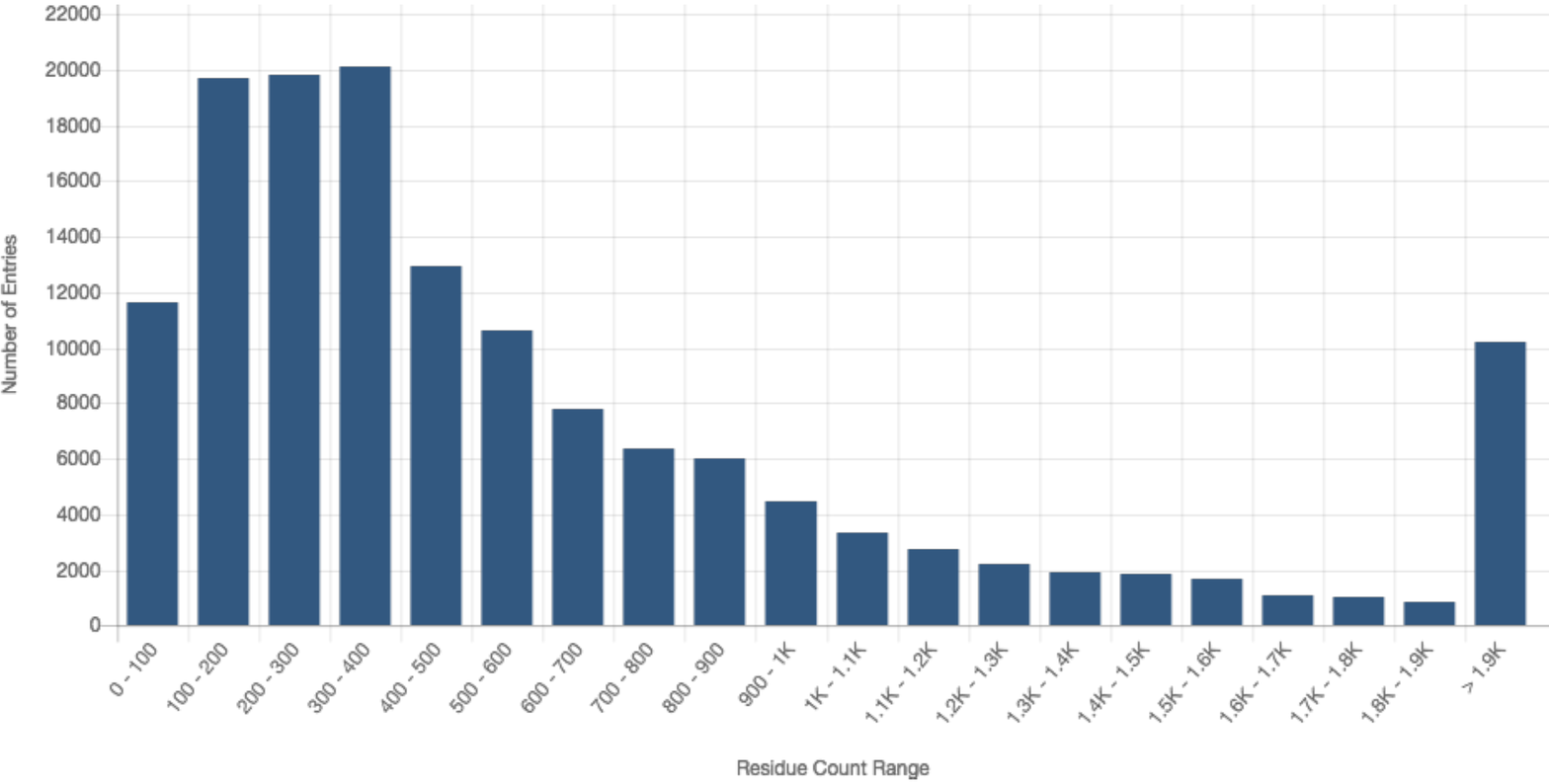} % second figure itself
       \vspace{-0.3in}
      \caption{Number of proteins in PDB by length.}
      \label{fig1}
  \end{minipage}
 
\end{figure}

The Protein Data Bank (PDB) is a very large and diverse dataset with proteins from different species and varying lengths and structures. Figure \ref{fig2} and Figure \ref{fig1} represent the variability of the dataset \cite{Berman00theprotein}.

% \begin{figure}[h]
% \label{fig1}
% \begin{center}
% \includegraphics[scale=0.3,angle=270]{graph_by_animal.pdf}
% \end{center}
% \end{figure}

% \begin{figure}[h]
% \label{fig2}
% \begin{center}
% \includegraphics[scale=0.3,angle=270]{graph_by_length.pdf}
% \end{center}
% \end{figure}

It is likely that proteins of drastically different sizes and proteins separated by billions of years of evolution rely on different sequence motifs for their structures. In fact, early tests in which we pulled proteins from the databank randomly fit very poorly. Thus, to address this challenge we created species-specific datasets for just two closely related animals, mouse and human, in order to increase similarity and patterning in the set. Due to the large size of the initial dataset, these subsets were still very large. 

We collected structural data of 30,000 human proteins and 6,000 mouse proteins.  We then subdivided each species-specific dataset, randomly assigning 80\% of our data as training data and 20\% of our data as test data. Many proteins are very symmetric including several sequence-identical chains.  In these cases, we used only one copy of each unique chain in order to reduce redundancy in the dataset.
 
\vspace{-0.1in} 
\subsection{Baselines}
 
\begin{figure}[ht]
\vspace{-0.1in}
\begin{center}
\includegraphics[width=0.4\textwidth, angle = 270, trim = 5.8cm 2cm 3cm 10.7cm, clip]{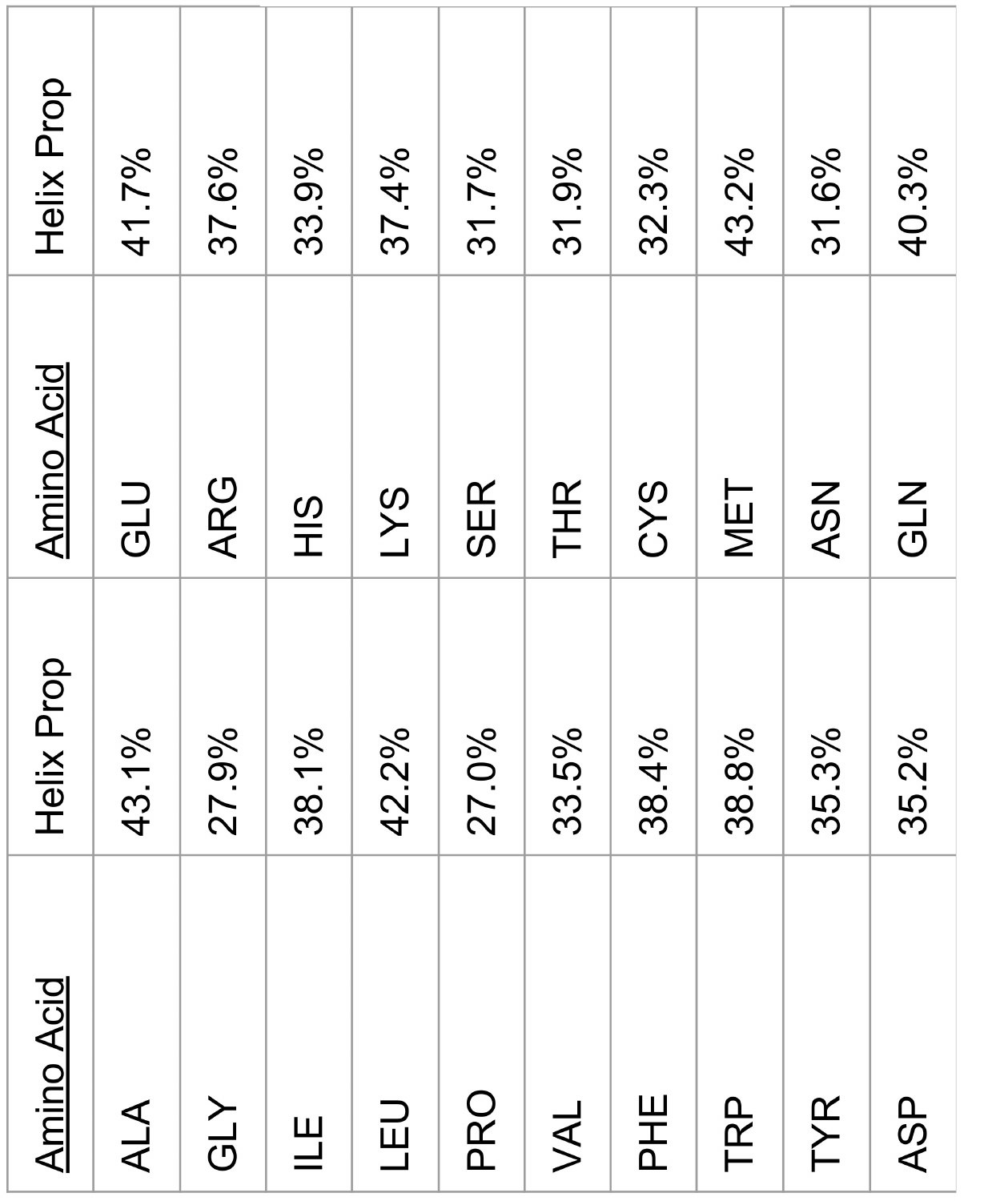}
\caption{Table of helix-probabilities for each acid.}
\label{fig-helixprod}
\end{center}
\vspace{-0.1in}
\end{figure}

We used a simple baseline against which we compared our more sophisticated models.  The helix-probability of a given amino acid type $R$ is defined
\begin{align*}
P(\text{Helix}| R) = H_R/N_R,
\end{align*}
where $N_R$ is the number of positions in the dataset which are of type $R$.  Let $H_R$ be the number of positions which are in a helix and of type $R$.  The helix probabilities for the 20 amino acids are shown in \autoref{fig-helixprod}.

This gives baseline probability $p_i$ that the position $i$ is part of helix given that it is of amino acid type $R$.  We can thus use the vector $(p_i)_i$ as a baseline prediction of helicity.  Using this baseline on the mouse dataset gives an average loss of $0.48$.    Any machine learning method which is not beating this rudimentary statistical approach is failing. 

A slightly more sophisticated baseline can be made by using sequence of 2 or 3 amino acid types such as
\[
P(\text{Helix}| R_1 R_2 R_3).
\]

Another basic score which can be used similarly is the helix-propensity of a given amino acid type $R$ defined
\[
P_R = \frac{P(\text{a position in a helix is type R})}{P(\text{a position is type R})}.
\]

%
%
%\[
%P_R = \frac{\text{frequency of $R$ in a helix in dataset}}{\text{frequency of $R$ in dataset}}.
%\] 
%\[
%P(\text{Helix}| R) = \frac{P(R|\text{Helix}) P(\text{Helix})}{P(R)}
%\]
%This can be used to predict the probability a position $i$ in a sequence is inside a helix given that the amino acid type $R$ appears at position $i$.    It can be estimated using our dataset.  Let $N$ be the number of amino acid positions in the dataset.  Let $H$ be the number of positions in a helix.  Let $N_R$ be the number of position in the dataset which are of type $R$.  Let $H_R$ be the number of positions which are in a helix and type $R$.    Then
%\begin{align*}
%P(\text{Helix}) &= H/N \\
%P(R) &= N_R/N \\
%P(R | \text{Helix}) &= H_R/ H \\
%\end{align*}
  
% \subsection{Simple Model-Random Forests} \label{rf}
% Before, implementing one hot encoding technique on our extracted protein sequence data from pdb files, 
% we tried directly mapping amino acids to their numbers (1-20) and then training them using Random
% Forests. We trained on an ensemble of 100 RFs with max height of 20, keeping all other parameters default.

%\subsection{Fully Connected Neural Network} 
 
\subsection{Model Implementation and Training} \label{NN}
%We trained the model trained using ensemble approach, and 

The implementation of our model can be found here: \url{https://github.com/sidharth0094/protein-structure-prediction}
%{https://github.com/sidharth0094/protein-structure-prediction}

Our FCNN trained on the mouse dataset had an average test loss of 0.21.  We include several graphs showing the outputs in \autoref{sec-appendix}.

We tried two different approaches to implement and train the FCNN presented in Section \ref{model}.
The first version of the model we implemented in Keras and trained by randomly shuffling protein sequence windows (each of size 10) in every epoch.    Whereas the second model we implemented in Pytorch and trained differently. In each epoch we shuffled the order of the proteins, but kept the sequence windows in order within each protein.

\begin{figure}[h]
  \centering
  \begin{minipage}{0.45\textwidth}
      \centering
\includegraphics[scale=0.22,angle=270,trim=6cm 6cm 6cm 6cm]{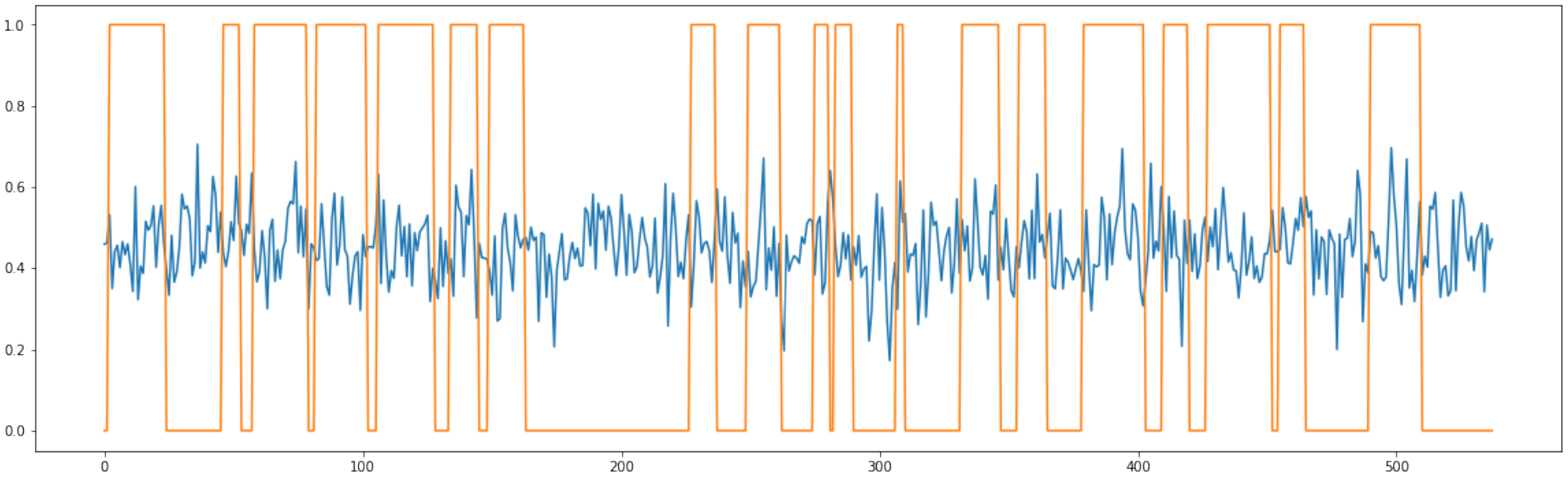}
\caption{First Training Run. }
      \label{fig-badtrain}
  \end{minipage}\hfill
  \begin{minipage}{0.45\textwidth}
      \centering
\includegraphics[scale=0.22,angle=270,trim=6cm 6cm 6cm 6cm]{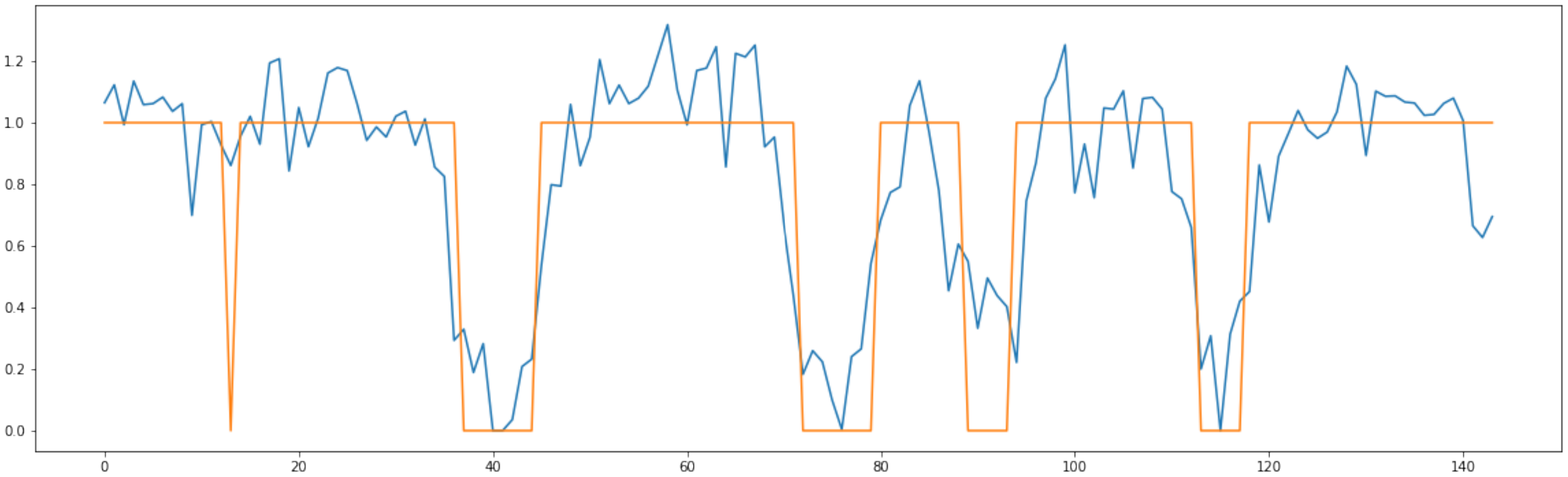}
\caption{Second Training Run. }
      \label{fig-goodtrain}
  \end{minipage}
  \caption{ The $x$-axis is the position index in the protein sequence.  The $y$-axis is probability of a helix (helicity).  The orange line shows the true presence of helices.  The blue line is the model's prediction of the probability of a helix.}
\end{figure}

The first model in Section \ref{NN} 
had a very high average loss of 0.6.  In \autoref{fig-badtrain}, we compare predicted helix probability to actual helix presence.  The prediction (blue line) is close to random noise around the mean.   However, the second model in Section \ref{NN}, predicted with average loss of 0.21 (\autoref{fig-goodtrain}).  

We conjecture the second model preformed better in part due the way in which we trained it.  In both neural network models, we randomly sampled proteins, and shuffled the order in each epoch. However, in the second model we ran through all the windows of each protein we selected \emph{in order}.  Thus, due to the overlapping windows, the model trained on each animo acid position several times, seeing it with slightly different context (surrounding sequence) each time. 

%\begin{figure}[h]
%\begin{center}
%\includegraphics[scale=0.3,angle=270,trim=6cm 6cm 6cm 6cm]{graph_bad_pdf}
%\caption{First Training Run.  The orange line shows the presence of helices within a given protein sequence.  The blue line is our models prediction of the probability of a helix.}
%\label{fig-badtrain}
%\end{center}
%\end{figure}
%\vspace{-0.1in}

%\begin{figure}[h]
%\begin{center}
%\includegraphics[scale=0.4,angle=270,trim=6cm 6cm 6cm 6cm]{graph_good_pdf}
%\caption{Second Training Run.  The orange line shows the presence of helices within a given protein sequence.  The blue line is our models prediction of the probability of a helix.}
%\label{fig-goodtrain}
%\end{center}
%\end{figure}
%\vspace{-0.1in}

\subsection{Cross Species Comparison}

\begin{figure}[ht]
\vspace{-0.1in}
\begin{center}
\includegraphics[scale=0.5,angle=0,trim=0cm 0cm 1cm 3cm, clip]{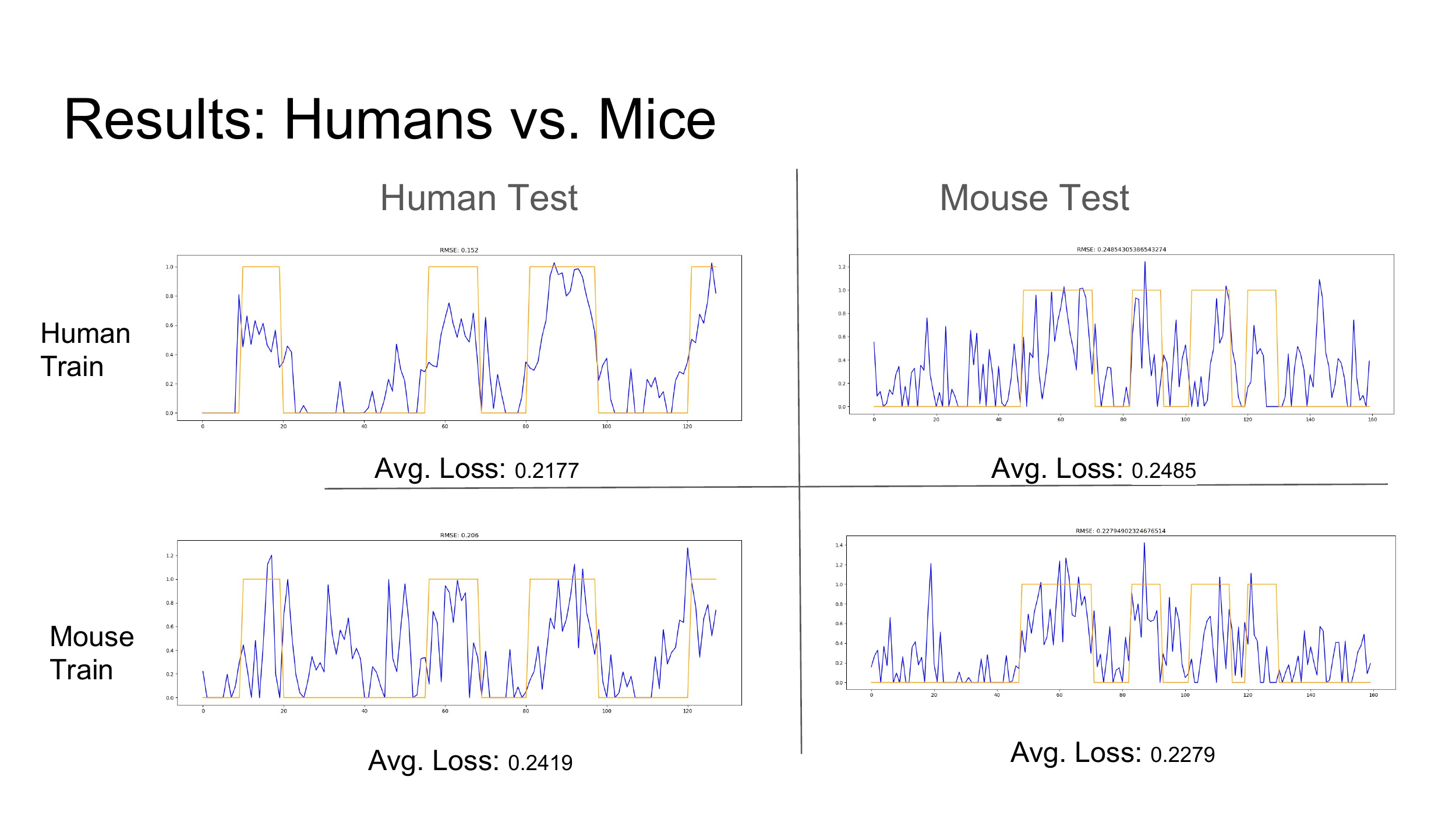}
\vspace{-0.1in}
\caption{Cross Species Comparison.  The Avg. Loss indicates the average RMSE over the entire test set, not just the example.}
\vspace{-0.2in}
\label{fig-humans-vs-mouse}
\end{center}

\end{figure}

We trained our FCNN model on random subsets of the mouse training set and human dataset of equal size (5000 proteins).  We then took each model and tested them on both their own species' test set and on the other species test set.  The results are illustrated in \autoref{fig-humans-vs-mouse}.  As expected, we found that each model fit its own species best.  The average RMSE for the human-trained human-tested model was 0.2177 and for mouse-trained mouse-tested it was 0.2279.  However, each model also did surprising well in the cross-species test.  The human-trained model had an average RMSE of 0.2485 on the mouse test set and mouse-trained model had an average RMSE of 0.2419 on the human test set.  

We interpret these results as a validation, in this context, of the use of mice as a model species for humans in laboratory environment.  The good cross-species fit implies that mouse and human proteins are largely very similar.  Looking at the losses for specific proteins shows that the mouse and human trained models often performed similarly on many targets but that the cross-species loss was worsened by certain more species-specific proteins which did not have strong analogues in the other training set.

\vspace{-0.1in}
\subsection{Error Function Comparison}
\vspace{-0.05in}

\begin{figure}[ht]
\begin{center}
\includegraphics[width=0.75\textwidth]{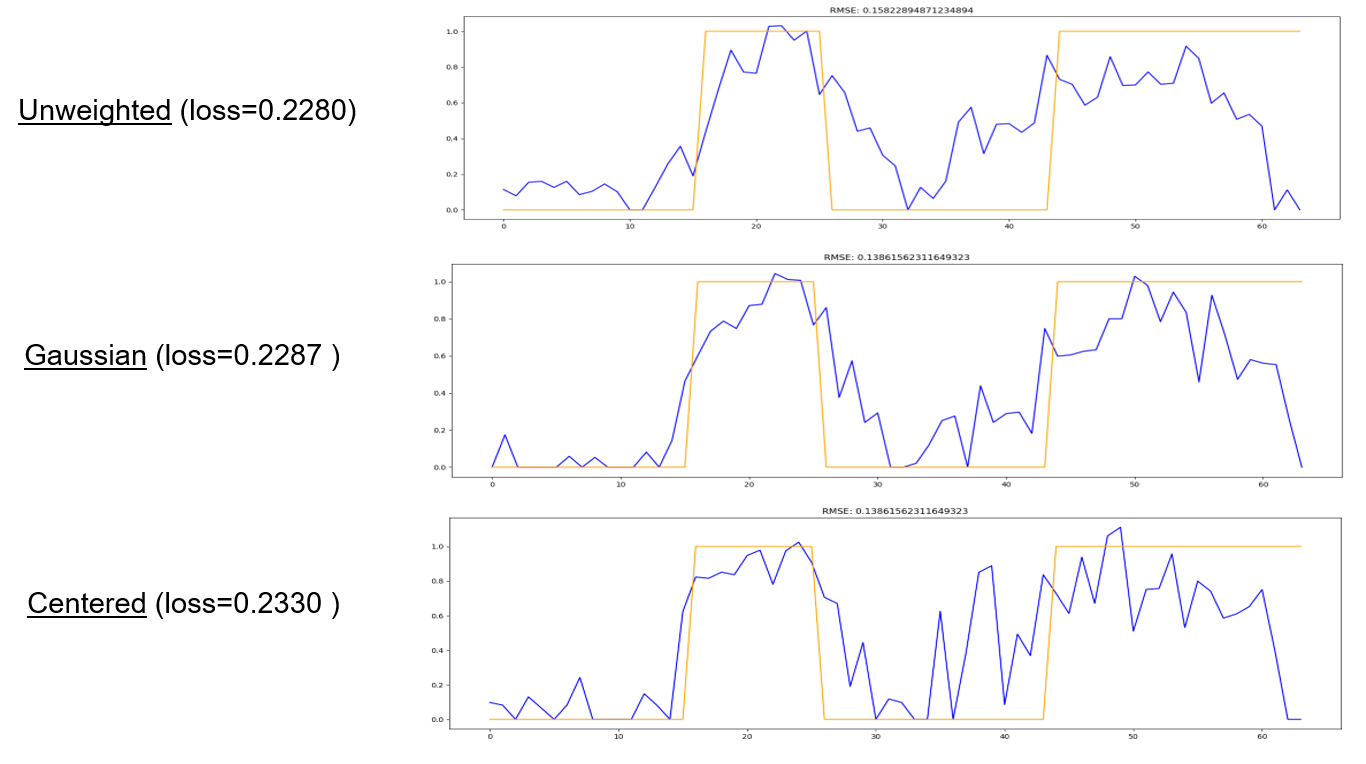}
\caption{Error Function Comparison.  The Avg. Loss indicates the average RMSE over the entire test set, not just the example.}
\label{fig-error_comp}
\end{center}
\end{figure}

We trained our FCNN model on the mouse training set using the different error functions mentioned in \autoref{sec-setup} and tested them on the mouse test set. The results are illustrated in \autoref{fig-error_comp}. Surprisingly, we found that the unweighted RMSE and Gaussian error performed almost the same with the average loss of 0.2280 and 0.2287 respectively. However, the centered error resulted in an average loss of 0.2330. The results from testing different error function contradicted our hypothesis that giving more weight to the amino acid at the center of the window would result in better prediction of the location of secondary structures.

\vspace{-0.05in}
\subsection{Window Size Comparison}
\vspace{-0.05in}

We conducted an experiment using our FCNN model on the mouse dataset with one hidden layer of 40 neurons and window sizes of 7, 10, 13. We selected these windows because each turn of the $\alpha$-helix takes 3.5 amino acids and thus these sizes correspond to 2, 3, and 4 turns respectively.  Our Hypothesis was that a window size of 10 or 13 should outperform the shorter window in predicting the accurate secondary structures since more information is available at longer lengths.

%We trained our MLP model on the mouse training set using window sizes of 7, 10, and 13.  One complete turn of an alpha helix corresponds to 3.6 amino acids, so these sizes are approximately 2,3, and 4 turns. 

\begin{figure}[ht]
\begin{center}
\includegraphics[width=0.75\textwidth]{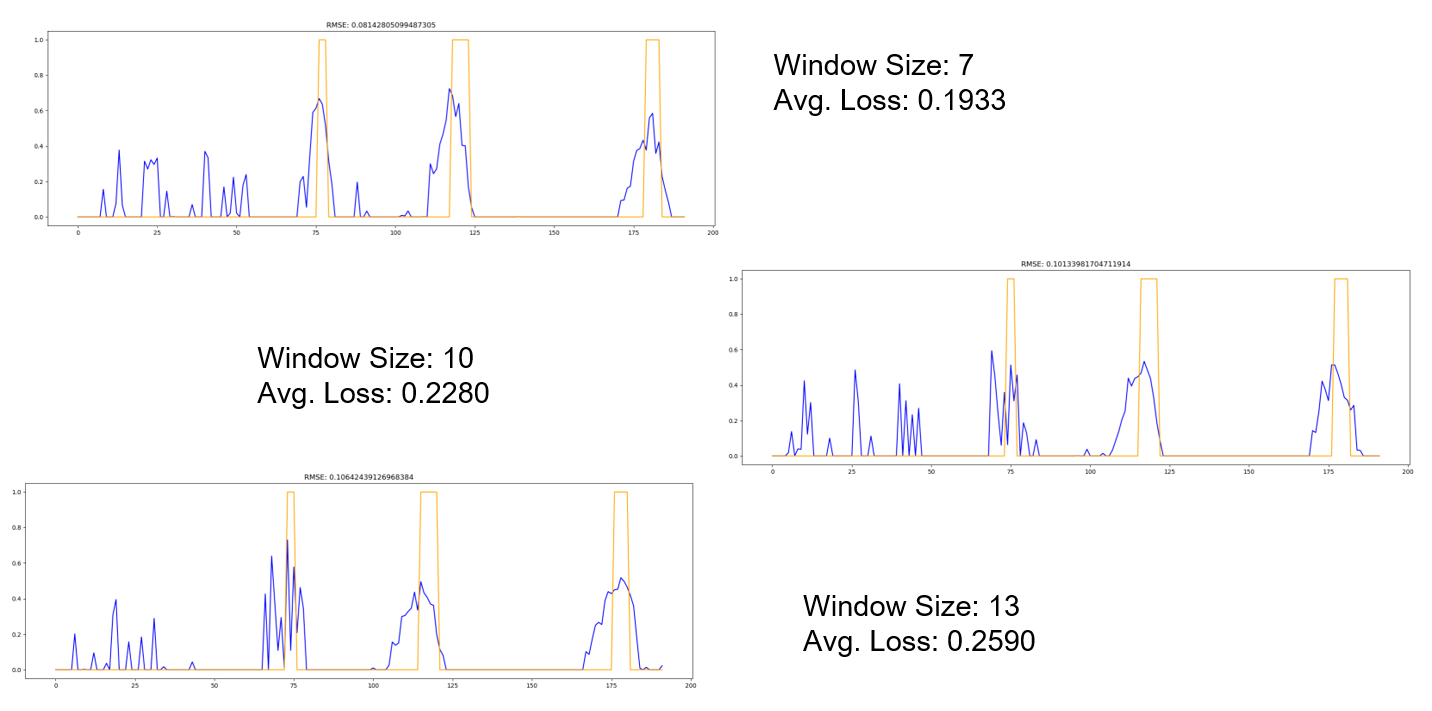}
\caption{Window Size Comparison.  The Avg. Loss indicates the average RMSE over the entire test set, not just the example.}
\label{fig-window_size_comp}
\end{center}
\end{figure}

The results are illustrated in \autoref{fig-window_size_comp}. The window size of 7 resulted in the average loss of 0.1933, whereas the window sizes of 10 and 13 resulted in average losses of 0.2280 and 0.2590 respectively.  These results contradicted our hypothesis that a window size of 10 or 13 would have a lower loss since larger windows sizes give the model greater context to predict structure.  Instead, it was the smallest window size which resulted in the best prediction.  It is likely 7 is close to the smallest size which would give good results.  A window size of 1, for example, implies no context and is essentially equivalent to using the baseline, which had average loss of $0.48$.

%ight as every turn in a helix structure in a protein contains 3.5 amino acids thus window of size 7 giving low average error as compared to other window sizes.

\section{Next Steps: Recurrent Neural Network}
\begin{figure}[h]
\begin{center}
\includegraphics[width=0.15\textwidth,angle=270, trim= 9cm 4cm 6cm 9cm, clip]{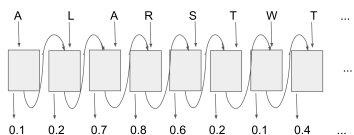}
\caption{Diagram of the RNN.}
\label{fig-RNN}
\end{center}
\end{figure}

We implemented a recurrent neural network which takes in a single amino acid at a time.  Let $\mathbf{x}^{(i)}$ be the length 20 encoding of the $i^{\mathrm{th}}$ amino acid.  Then define $h^{(0)} = \mathbf{0}$ and $h^{(k)}_i = R\left(\sum_{j=1}^{20} w^{[1]}_{i,j} x^{(k)}_j + \sum_{j=1}^{40} w^{[1]}_{i,20+j} h^{(k-1)}_j \right)$ and $\mathbf{o}^{(k)}_i = R\left( \sum_{j=1}^{40} w^{[2]}_{i,j} h^{(k)}_j \right)$.  We can visualize this network as in \autoref{fig-RNN}.

We wish to train this recurrent neural network and compare the error to the FCNN. Since, protein chains are sequential in nature, we believe that an RNN should be able to capture the patterns in the dataset better than a FCNN.

\bibliographystyle{acm}
\bibliography{refs}

\appendix

\section{Example Predictions from the Mouse FCNN}

\label{sec-appendix}

We present some illustrations of the predictions made by the FCNN on the mouse test set.  In each graph, the $x$-axis is the position index in the protein sequence.  The $y$-axis is probability of a helix.  The orange line shows the true presence of helices.  The blue line is the models prediction of the probability of a helix.

\begin{figure}[ht]
\begin{center}
\includegraphics[width=0.25\textwidth,angle=270, trim= 11.3cm 5cm 4cm 3.5cm, clip]{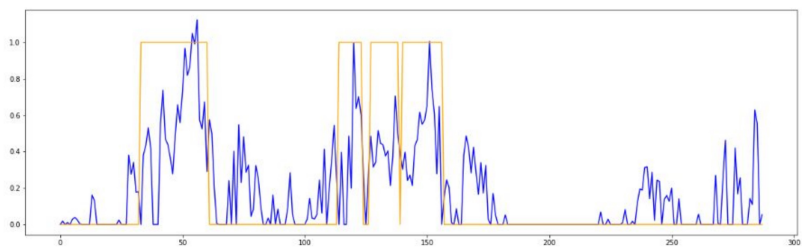}
\caption{Protein 1CD1.  Loss = 0.15.  A very good result.  Note the very strong dip in the blue line between the second and third helix.}
\end{center}
\end{figure}

\begin{figure}[ht]
\vspace{-0.1in}
\begin{center}
\includegraphics[width=1\textwidth]{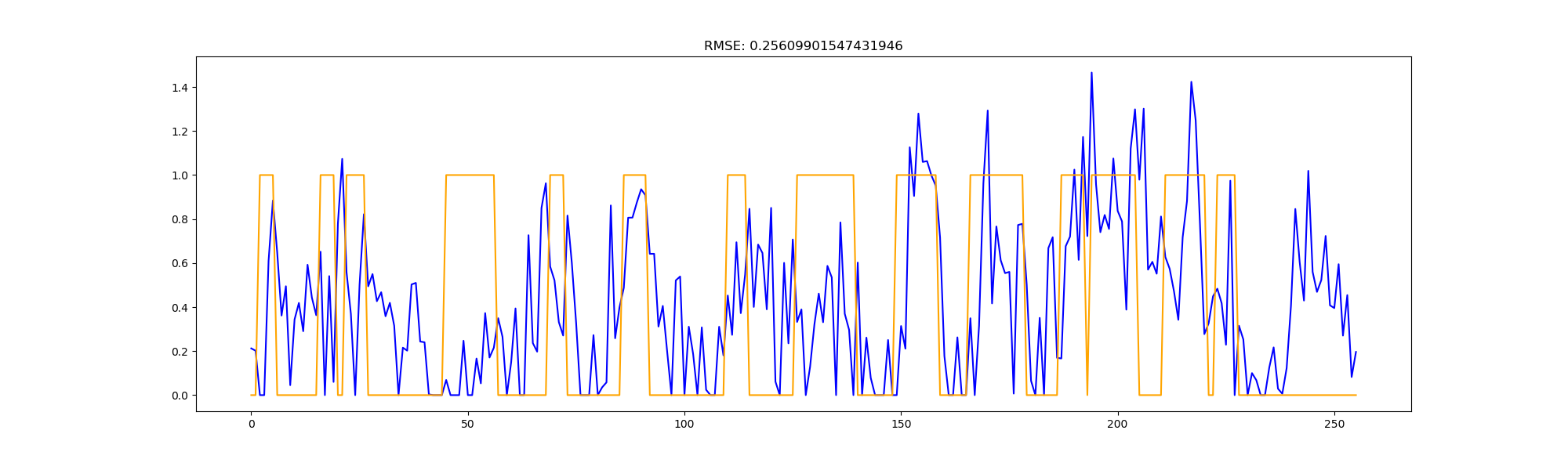}
\vspace{-0.1in}
\caption{Protein 1AQU. Loss = 0.256. A medium-quality result.  Many helices are clearly marked by spikes, but the beginning and ending points are unclear.  Some helices are missed outright and small valleys between helices are indistinct.}
\vspace{-0.2in}
\end{center}
\end{figure}

\begin{figure}[ht]
\vspace{-0.1in}
\begin{center}
\includegraphics[width=1\textwidth]{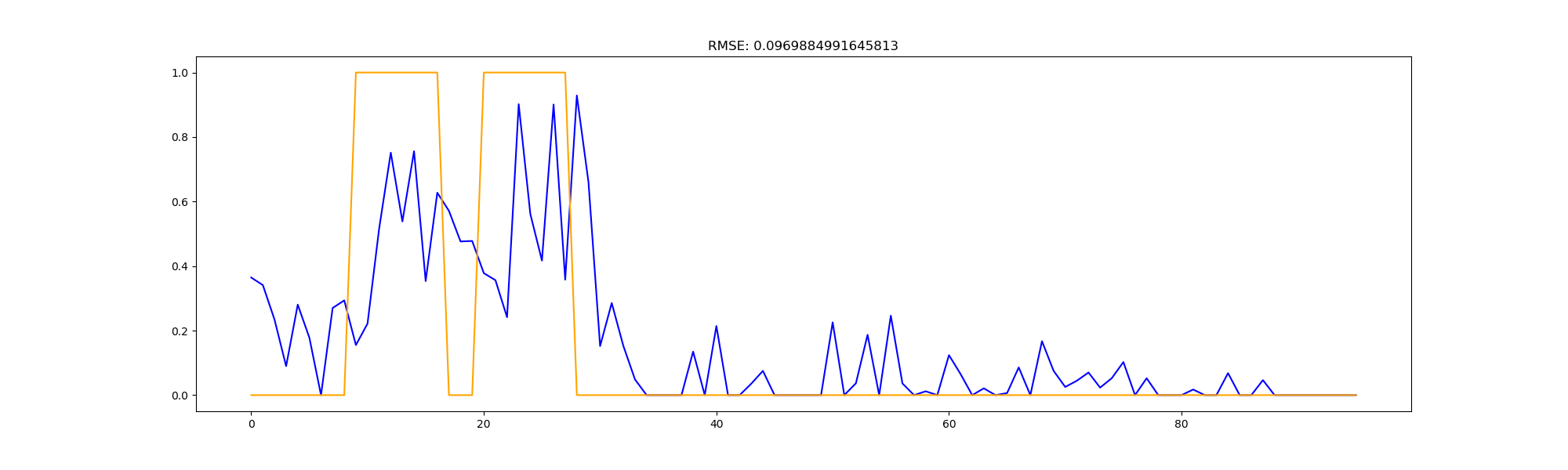}
\vspace{-0.1in}
\caption{Protein 1AB0. Loss = 0.097. A good result.  Clearly detects both helices and their lengths but errs in predicting their positions slightly too late in the sequence.  This result also does a good job of predicting no helices through the rest of the sequence. }
\vspace{-0.2in}
\end{center}
\end{figure}

% \begin{figure}[ht]
% \vspace{-0.1in}
% \begin{center}
% \includegraphics[width=1\textwidth]{1cyd.png}
% \vspace{-0.1in}
% \caption{Protein 1CYD. Loss = 0.2824}
% \vspace{-0.2in}
% \end{center}
% \end{figure}

\begin{figure}[ht]
\vspace{-0.1in}
\begin{center}
\includegraphics[width=0.8\textwidth]{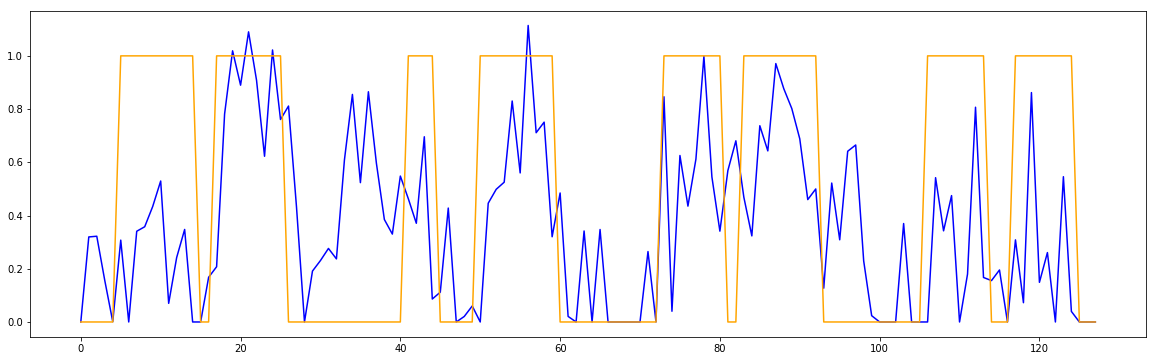}
\vspace{-0.1in}
\caption{Protein 1AP7.  Loss = 0.2462}
\vspace{-0.2in}
\end{center}
\end{figure}

\begin{figure}[!t]
\vspace{-0.1in}
\begin{center}
\includegraphics[width=1\textwidth]{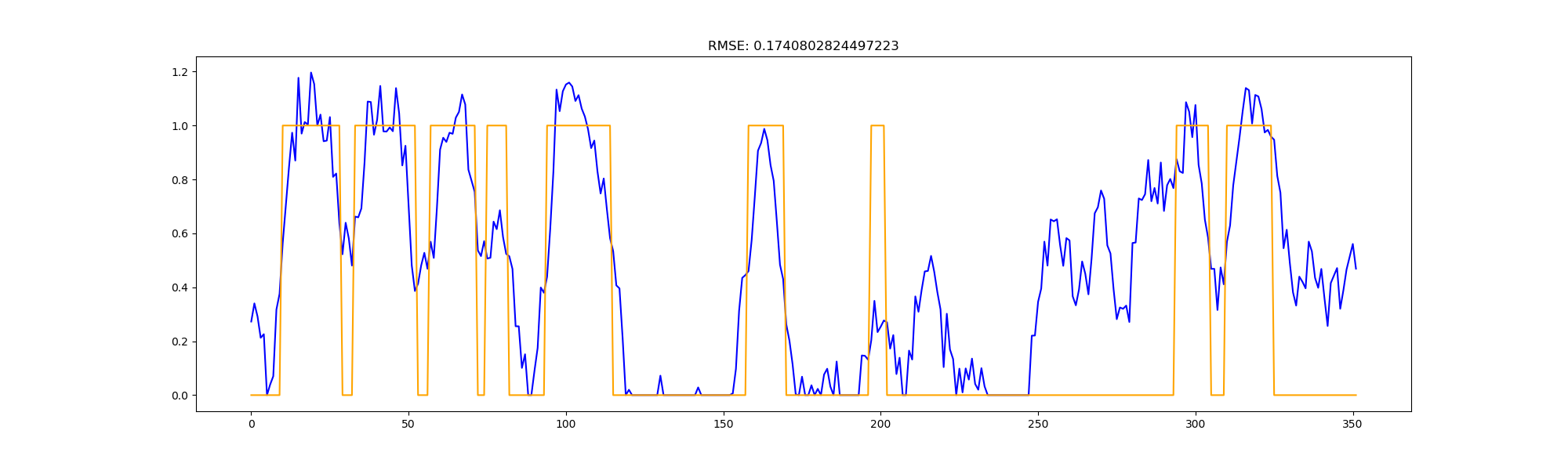}
\vspace{-0.1in}
\caption{Protein 1DD7.  Loss = 0.1741}
\vspace{-0.2in}
\end{center}
\end{figure}

\end{document}